# AGI Maze as a Benchmark Framework for World-Modeling Agents

Alexey Potapov

SingularityNET Foundation
alexey@singularitynet.io

## Abstract

Large language models (LLMs) are powerful pattern-completion systems, but their default operating mode – predicting the next token from a static context – does not reliably produce persistent, manipulable representations of an external world. Many tasks that look like "reasoning" in text become substantially harder once the environment is **partially observable**, **stateful**, and requires **memory** and **structured hypotheses** about hidden state.

AGI Maze is a lightweight framework for building such environments without requiring high-dimensional sensory inputs. It provides a family of grid-based maze tasks with a clean API and multiple difficulty regimes. The goal is to create benchmarks where agents must learn and use **world state representations**, not just infer a local rule over readily provided observations.

We provide an initial evaluation of several vanilla LLMs on simple mazes showing that they fail to represent mazes internally at LLM inference time. We also introduce a baseline agent, which is allowed to use its message history as a working memory to construct descriptions of observations at agentic runtime. Although this can improve performance, it is still insufficient for an LLM agent to reliably solve even small mazes within a step budget that is more than enough for humans.

## 1. Motivation: from LLM limitations to world models

### 1.1 LLMs are static predictors, not persistent world simulators

A core limitation of current LLM-only agents is that the model itself is *static*: the only "dynamics" typically come from:

- adding text to the prompt (message history as working memory),
- retrieving external text (RAG) into the prompt (long-term memory),
- and feeding tool outputs back as text.

This creates two intertwined problems:

1. **Memory as text is inefficient.** Some tasks require explicit state tracking, revisiting earlier observations, and performing multi-step search. Encoding everything as unstructured text leads to brittle behavior and high token cost.
2. **Representation is not guaranteed.** Next-token prediction encourages extracting whatever information is needed to produce the next output, not maintaining a stable, manipulable representation of “what is true in the world.” In language tasks this deficiency is partially hidden because language already *is* a world description; in interactive environments, it becomes obvious.

We discussed how the architecture and the training objective of LLMs are connected with their practical limitation in the paper [1] in detail.

A complementary angle comes from empirical observations that internal activations can encode much of the final output early in the network, which supports the view that many models are best described as efficient predictors rather than explicit world-state modelers. See e.g. [2] (the discussion around intermediate-layer information content including possibility to remove, swap and iterate over blocks of transformers in LLMs, including references [3-6]).

## 1.2 Why “world models” need more than rule discovery

In the language domain, "world modeling" is often reduced to "discover the rule of a game." But for agents, a full notion of world modeling also includes:

- representing **latent state** (what you cannot currently observe),
- maintaining **beliefs** under uncertainty,
- updating those beliefs with new evidence,
- reasoning over representations (maps, graphs, causal schemas),
- and using memory efficiently (working memory, episodic memory, long-term knowledge).

In other words: world models are about **state and representation**, not only about rules.

## 1.3 World modeling testbeds and their limitations

AGI-ARC-3 [7] is an important benchmark for *generalization across tasks* and for inferring hidden rules. However, many ARC-like settings:

- are effectively **fully observable**,
- do not require agents to maintain a persistent, queryable **world state** across time,
- and do not pressure-test partial observability, localization, mapping, and long-horizon memory.

AGI-ARC-3 targets an important slice of AGI, but it is not a direct test of "can the agent construct and use a representation of an evolving world state?" Indeed, this challenge can be solved to a large extent [8] by LLM agents, which construct compact executable world models externally. We believe that this is insufficient for AGI.

This insufficiency can be more clearly seen in the domain of embodied agents. Direct extensions of LLMs to this domain, namely vision-language-action models, can learn to perform short sequences of actions. Using RAGs over vector embeddings of video frames, they can recall visual scenes of previously encountered relevant places, but they do not solve the simultaneous localization and mapping problem (do not construct a world state representation) to navigate back to such places when needed (e.g., see capabilities of VLA agents in Minecraft [9]). This is why the world modeling problem is frequently stated in visually observable environments (e.g., [10]).

However, here, world models are considered in quite a simplistic way (quite close to visual features at the moment).

In this paper, we introduce AGI Maze as a framework for benchmarking world modeling capabilities of AI agents in the language domain. This framework avoids complexities of low-level representation learning, but requires an agent to reconstruct descriptions of environments under partial observability and uncertainty, and to reason over them.

# 2. AGI Maze: a lightweight testbed for state, memory, and representation

## 2.1 Design goals

AGI Maze aims to be:

- **interactive and stateful**, but with low bandwidth (no pixels required),
- **partially observable** by default (except tutorial mode),
- simple enough for humans to play,
- extensible to new mechanics,
- and usable via a stable HTTP API.

## 2.2 The base environment: grid + walls + monolith + exit

The agent lives on a grid (most typical sizes range from 3x3 to 8x8). Each step it chooses an action from:

- up / down / left / right

The maze has:

- internal walls,
- an unbreakable border ("monolith") with exactly one opening (the exit),
- a treasure chest and its key.

Key constraints:

- stepping into a wall/monolith does not change position but consumes a step,
- treasure requires the key,
- exiting requires the treasure,
- some instances also require an exit key.

Figure 1 shows an example of such a maze.

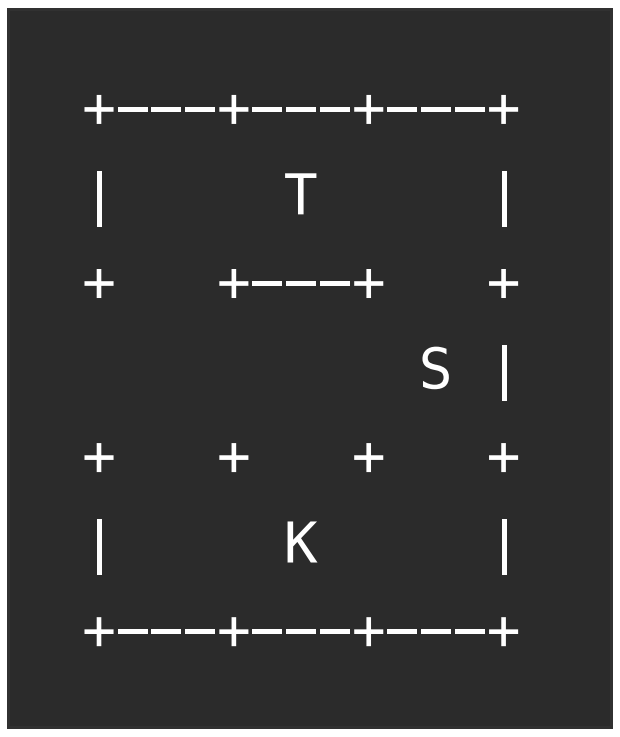

Figure 1. 3x3 maze example: S - start, K - key, T - treasure

The maze is not shown to the player. Only the start position and the grid size are provided. Observations are provided textually (e.g., "You tried to go right, but a monolith blocks the way"). Thus, the agent by itself should construct the map or somehow else track its location, known walls and passages, and unvisited cells.

However, this alone doesn't yet create a dynamically ambiguous POMDP-like situation, in which the same observation can correspond to multiple locations, and the agent must maintain a belief (not precise knowledge) over where it might be. The environment becomes interesting and challenging, when the following elements are added to the maze.

## 2.3 Dynamics that amplify partial observability

To make state inference and representation genuinely necessary, AGI Maze includes mechanics such as:

- **Rivers**: stepping onto a river cell triggers forced downstream movement (e.g., 2 cells, or 1 near the mouth), but the agent is not informed about the river's flow direction. The agent must reason about both the intentional move and the forced transition.
- **Pit cycles**: stepping into a pit teleports the agent to the next pit in a cycle (wrapping around). Multiple cycles may exist.

These mechanics are deliberately chosen because naive "grid exploration" strategies break down: localization and mapping require explicit hypotheses.

It should also be noted that the number of allowed steps is limited (the threshold depends on the maze size and difficulty). While such mazes are supposed to be used in benchmarking, in which no critical mistake can be made due to partial observability, the step budget requires the agent to navigate and explore cleverly. In difficult mazes, crucial elements such as chests, can be placed on such cells, which can be reached only with a strict sequence of actions.

Figure 2 shows the maze, in which the key can only be reached by jumping into the river in the upper left corner and then moving left two times. Reaching the exit requires jumping into the second river source and then moving right and down. The agent should not only reconstruct the map from incomplete observations, but also find the necessary sequence of actions by reasoning, otherwise it fails to meet the step budget.

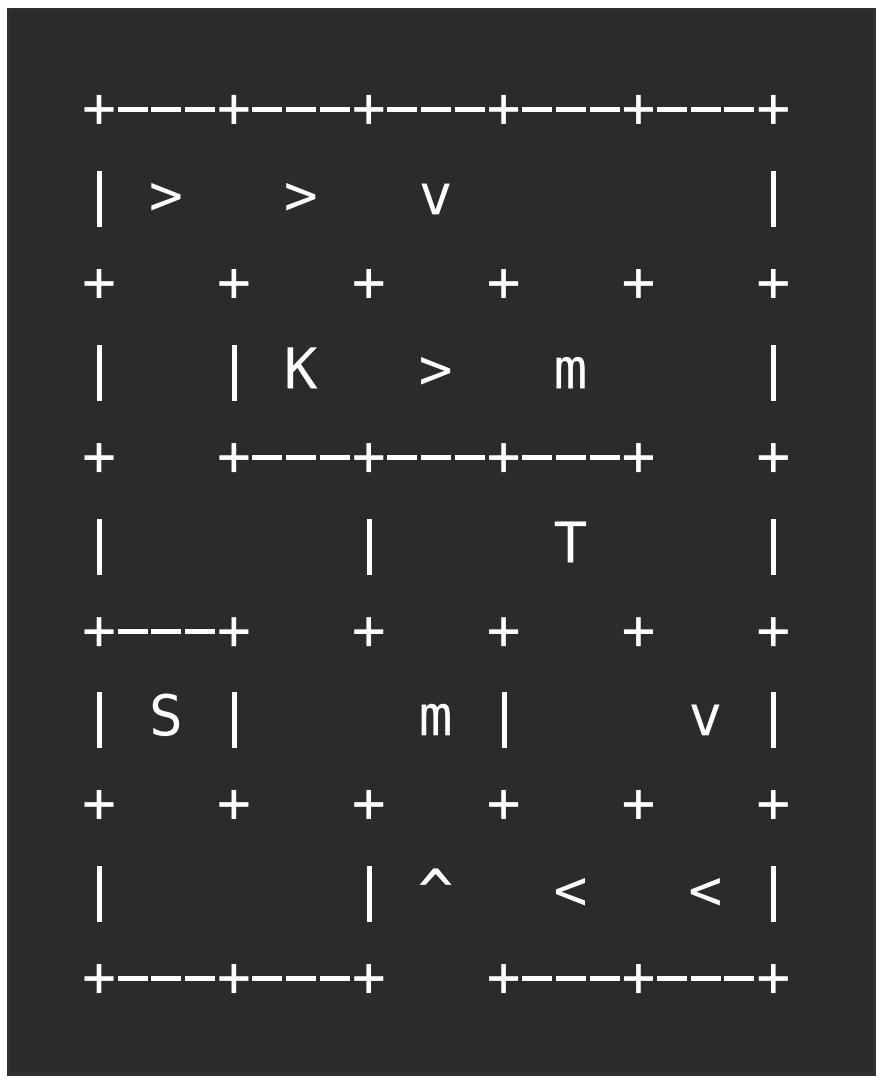

Figure 2. A difficult maze (<, V, >, ^ – river cells, m – river mouth)

## 2.4 Extensions as a generality test

The core rules are enough to create challenging mazes, which require inventing non-trivial heuristics and strategies, but it is difficult to prevent researchers from hand-coding certain tools or engineering prompts, which will greatly help agents to solve these mazes. We are proposing the AGI Maze framework not as an ultimate AGI benchmark, but rather as guidance for researching what is missing in LLM (or other) agents. And we don't want to make subjective judgements whether agents being tested are AGI-ish enough or overfit.

To motivate researchers to avoid handcrafting and overfitting, we make the framework to support extensions beyond the core rules that introduce:

- new items (e.g., boat, grenades, flashlight),
- new cell types,
- new actions (e.g., blast, shine).

Crucially, new elements require new heuristics and strategies in order to meet the step budget. For example, introduction of the flashlight (that provides information about a neighbor cell without moving to it) requires the agent to figure out, in which situations its usage will help save moves or reducing uncertainty.

Moreover, some extensions can create situations where reaching the goal is completely impossible without understanding the new mechanic (e.g., an exit behind a river that is only crossable with a boat; a key sealed behind a blastable wall).

This makes "general world understanding" (in a narrow but meaningful sense) observable: does the agent infer new mechanics (or/and navigation and exploration strategies for it) from interaction, or does it rely on brittle hardcoding?

# 3. The framework structure

## 3.1 Task groups and what they are for

Currently, mazes are divided into 5 large groups.

- **TUTORIAL**: open map; for teaching humans/agents the basic mechanics; not used for scoring.
- **TRAINING**: small mazes + generous step budgets; useful for iteration, calibration, RL training, and baseline benchmarking.
- **CLASSIC**: larger mazes under core rules; calibrated against humans; passing indicates a strong solver.
- **EXTENDED**: tests generality via new mechanics.
- **HIDDEN**: private hold-out; validates that EXTENDED performance is not merely dataset-fitting. It contains extensions with new items and cell types, which are not known to researchers, and which mechanics may or may not be described to agents. In the latter case, the agent should infer underlying rules from interactions.

Let us note that we do not introduce a particular benchmark here, but propose the framework, which can be used to construct different benchmarks depending on various research goals. However, the presence of **HIDDEN** highlights the ultimate goal: not to "beat the benchmark" at any cost, but to demonstrate general world-modeling capabilities that transfer.

## 3.2 API

The AGI Maze website[1] provides a human playing mode as well as an API for agents. This API allows retrieving the game description (GET /api/description), which is identical to the description given to humans (and thus recommended for using in agents).

Starting a new game (POST /api/start) allows to choose a maze, and returns the same information about the chosen maze as humans receive in Web UI (maze sizes, starting position, available actions, inventory). This information should be provided to the agent. The response also contains the recommended step budget.

Then, the agent makes steps (POST /api/step) sending the selected action until the maze is solved. The response contains observations in textual form (the same as for humans) as well as in a more structured format, current inventory, and some other fields. The game engine doesn't prevent continuing playing after the step budget is exceeded. The detailed API description and examples of agents are provided in the agimaze-bench repo[2].

---

[1] https://agimaze.org/

[2] https://github.com/Necr0x0Der/agimaze-bench

# 4. Baseline benchmark example

## 4.1 Vanilla LLM agents

In this section, we introduce a basic example of benchmarking within the proposed framework. We study capabilities of vanilla LLM agents to solve basic mazes. The vanilla LLM agent, which directly executes an LLM on the list of the following input messages:

- the general game description (obtained via API);
- the current maze parameters (obtained in response to the new game start);
- the sequence of actions performed by the agent and textual observations received in response to them.

The agent is provided with one tool, which actions correspond to the available game actions, and is prompted to use it. Thus, an LLM should infer its next action directly from the history of action-observation pairs prepended by the game rules. This agent doesn't construct a map or anything else as an external artifact, and doesn't have any kind of memory beyond the raw history in the messages / prompt. Thus, capabilities of vanilla LLMs are being tested here.

We performed massive testing of light LLMs only against a random-walk agent (which simply randomly chooses from four actions at each step). None of them appeared to be capable of solving even simple mazes reliably from **TRAINING**, so we didn't test them on other groups of mazes. Table 1 summarizes the results.

Table 1. Success rates (%) of the vanilla LLM agent (random – random-walk agent; gpt4om – GPT-4o Mini, gpt5m – GPT-5 Mini, gem31fl – Gemini 3.1 Flash Light, minimax27 – MiniMax M2.7)

| maze | size | step budget | random | gpt4om | gpt5m | gem31fl | minimax27 |
|---|---|---|---|---|---|---|---|
| S0-01 | 3x3 | 70 (35) | 24.3 (4.5) | 11 (4) | 65 (37) | 79 (43) | 28 (11) |
| S0-02 | 3x3 | 70 (35) | 16.7 (2.2) | 4 (1) | 49 (26) | 71 (40) | 28 (9) |
| S0-03 | 4x4 | 160 (80) | 9.2 (1.3) | 4 (0) | 8 (3) | 39 (13) | 17 (6) |
| S1-01 | 3x3 | 70 (35) | 22.3 (4.3) | 2 (0) | 53 (36) | 51 (17) | 8 (3) |
| S2-01 | 3x3 | 70 (35) | 24.8 (4.8) | 23 (6) | 36 (15) | 42 (10) | 34 (8) |
| S3-01 | 4x3 | 100 (50) | 20.6 (4.5) | 10 (0) | 25 (4) | 27 (15) | 9 (3) |
| S3-03 | 5x4 | 200 (100) | 3.2 (0.4) | 0 | 3 (1) | 1 (0) | 0 |

Each maze entry (S0-01, ..., S3-03) contains 10 mazes of the same size and rules (10 trials per each maze were performed for each LLM, so each number is the number of successful trials out of 100).

There are only keys and chests in S0. There are rivers in S1 and pits in S2. Mazes from S3 contain all the elements (pits and rivers simultaneously in addition to keys and chests).

We used two thresholds on the number of steps. Calibration against human performance is supposed to be done for **CLASSIC**, **EXTENDED**, and **HIDDEN**, but let us note that the lower threshold is soft but reasonable for human players. Thus, if the agents do not show 100% score on the **TRAINING** with the lower step budget, they are quite weak. The upper threshold is twice the lower threshold.

It can be seen that the tested LLMs do not achieve 100% score not only with the lower step budget, but with the higher budget as well. In fact, GPT-4o Mini appeared to be systematically weaker than the random-walk agent, while others are typically stronger than random on small mazes. Gemini-3.1-Flash-Light showed the best result on small mazes especially without rivers and pits, but none of these models outperformed the random-walk agent for mazes 5x4 and larger.

One can derive some conclusions about strengths of particular LLMs and weaknesses of LLMs in general analyzing relative performance of LLMs on different sizes or comparing it on mazes with and without dynamic uncertainty. However, generally speaking, we can suppose that LLMs manage to do some spatial inference internally, but its capabilities are very limited.

Of course, one can ask whether stronger models can do better. We performed a few runs on S3-03 and observed about 30% success rate with doubled step budget for GPT-5.5 and 40% for Gemini 3.5 Flash, which is considerably better than that of lighter LLMs, but which is incomparable to humans.

## 4.2 Prompt as paper and pencil

It would be fair to say that humans can reliably solve only relatively simple mazes without a pencil and a piece of paper, and vanilla LLMs are deprived of this. Thus, allowing LLM agents to make notes in their prompts would be a natural next step. However, let us note that vanilla LLMs are weaker in solving AGI Maze than humans even without a notepad.

In order to see to what extent using notes in the prompt to maintain the environment state and plans can help LLMs solve mazes, we designed a simple planning agent. This agent performs each action in two stages:

- first, it is prompted to make notes to itself about the current exploration status and future actions considering these notes as working memory;
- second, the agent is prompted to make a tool call to perform an action.

Typical entries of this working memory look like (of course, they can be different for different LLMs or if different instructions of how to use the working memory is given):

- "Moving down from (1,2) to (2,2) to explore the bottom-right corner and continue mapping the maze to find the key."
- "Moving right from (2,1) to (2,2) to head toward the chest at (1,2) to collect the treasure." (after the key is found)

- "I'm at (2,0) on the bottom border. All 9 tiles have been explored and appear empty, but the exit opening must be one of the border cells. Trying to move down from (2,0) to exit the maze through the monolith opening."

It can be seen that providing an LLM a possibility to use the message history as a working memory with minimalistic instructions allows an agent to keep track of its coordinate, the state of exploration, coordinates of the chest, future actions, etc. Although the agent doesn't try to construct the whole map in this case. It can be interesting to see, how different LLMs reason about pits and rivers (when coordinates become unknown), what prompts without domain-specific hints can help them to construct more reliable state descriptions, and so on, but this goes outside the scope of our paper. But we report the following improvements on S3-03 for switching from the vanilla LLM agent to this planning agent (for doubled step budget):

- GPT 5.5 : 30% → 60%
- Gemini 3.5 Flash : 40% → 70%

This increase indicates that this information, which LLMs put into their notes, is not constructed by LLMs in internal latent representations at intermediate layers within the inference process over plain interaction history (although this information directly follows from this history).

Interestingly, while there is a considerable improvement for GPT 5.5 and Gemini 3.5 Flash, the planning agent with lighter LLMs (listed in Table 1) showed zero score (no apparent improvement) on S3-03.

# Conclusion

AGI Maze is a lightweight framework for studying and benchmarking **world-modeling** in the sense that matters for agents: constructing and using representations of a **latent, evolving world state** under partial observability, and reasoning about that state over long horizons under a step budget.

The baseline results reported here support a simple, but important, qualitative picture. Vanilla LLM agents — given the same English observations as humans — often fail to solve even small **TRAINING** mazes reliably, and some models can be systematically worse than a random-walk baseline. This is consistent with the hypothesis that, during standard next-token inference, LLMs do not reliably build an explicit latent representation of the environment state that they can update and query as a coherent map.

Allowing an agent to use the prompt/message history as a form of **working memory** (e.g., by writing notes before acting) can improve performance for stronger models. However, the improvements are limited: without explicit instructions to externalize state, LLMs typically do not attempt to build a full map in their notes, and the resulting "memory-as-text" approach remains brittle and inefficient.

From a benchmark-design perspective, AGI Maze intentionally pressures the capabilities that this gap exposes:

- belief tracking under partial observability,

- map construction (or alternative state representations),
- memory use and compression (what to store, how to update it),
- long-horizon planning under uncertainty,
- and adaptation to new mechanics via extensions.

At the same time, the framework deliberately avoids pixel-level perception and continuous control. The point is not that perception is unimportant, but that removing it helps isolate representation and reasoning as the primary bottleneck.

A natural next step is to evaluate LLM agents that build **explicit, computable world models** as external artifacts (e.g., maps, graphs, constraint systems, or programmatic simulators). We expect such approaches to significantly improve results, but also hypothesize that they will not be sufficient by themselves: robust world modeling will likely require better mechanisms for uncertainty handling, memory management, and strategy learning beyond ad hoc tooling.

Finally, AGI Maze should be viewed as a **framework**, not a single fixed benchmark. While we use it here to study LLM-based agents and world-modeling failures, the same task families and API can support other research directions as well — for example, training and evaluating RL agents, studying exploration under uncertainty, or testing different memory architectures.